\documentclass[compsoc]{IEEEtran}
\def\IEEEcompsocdiamondline{}
\IEEEoverridecommandlockouts
\renewcommand\IEEEkeywordsname{\bf Keywords}

\makeatletter
\renewcommand\footnoterule{%
\kern-3\p@
\hrule\@width.4\columnwidth
\kern2.6\p@}
\makeatother

\usepackage{hyperref}

\usepackage{academicons}
\usepackage{amsmath}
\usepackage{amssymb,amsfonts}
\usepackage{algorithmic}
\usepackage{graphicx}
\usepackage{textcomp}
\usepackage{float}
\usepackage{xcolor}
\usepackage{tikz}
\usetikzlibrary{shapes,arrows.meta,decorations.pathmorphing,backgrounds,positioning,fit,petri,calc,hobby,external}
\usepackage{cleveref}
\usepackage{caption}
\usepackage{boldline,multirow}
\usepackage{booktabs}
\usepackage{etoolbox,siunitx}
\usepackage{array}
\newcolumntype{P}[1]{>{\centering\arraybackslash}p{#1}}
\usepackage{adjustbox}
\usepackage{datetime2}

\tikzset{%
  every neuron/.style={
    circle,
    draw,
    minimum size=0.5cm
  },
  neuron missing/.style={
    draw=none, 
    scale=2,
    text height=0.333cm,
    execute at begin node=$\vdots$
  },
}

\tikzstyle{block} = [rectangle, draw, fill = blue!20, text width = 8em, text centered, rounded corners, inner sep = 8pt, minimum height = 4em]
\tikzstyle{wideblock} = [rectangle, draw, fill = blue!20, text width = 13em, text centered, rounded corners, inner sep = 11pt, minimum height = 4em]
\tikzstyle{smallblock} = [rectangle, draw, fill = blue!20, text width = 5em, text centered, rounded corners, inner sep = 4pt, minimum height = 3em]
\tikzstyle{emptyblock} = [rectangle, text width = 8em, text centered, minimum height = 4em]
\tikzstyle{smallcircleblock} = [circle, draw, text width = 0.7em, text centered]

\newcommand{\orcid}[1]{\href{https://orcid.org/#1}{\textcolor[HTML]{A6CE39}{\aiOrcid}}}

\usepackage[left=1.77cm,right=1.77cm,top=1.55cm,bottom=2.35cm]{geometry}

\usepackage{listings}
\usepackage{color}

\definecolor{codegreen}{rgb}{0,0.6,0}
\definecolor{codegray}{rgb}{0.5,0.5,0.5}
\definecolor{codepurple}{rgb}{0.58,0,0.82}
\definecolor{backcolour}{RGB}{242,242,235}

\usepackage[style=ieee,natbib=true]{biblatex}
\renewrobustcmd{\bfseries}{\fontseries{b}\selectfont}
\renewrobustcmd{\boldmath}{}
\newrobustcmd{\B}{\bfseries}

\def\BibTeX{{\rm B\kern-.05em{\sc i\kern-.025em b}\kern-.08em
    T\kern-.1667em\lower.7ex\hbox{E}\kern-.125emX}}

\begin{document}
\bstctlcite{IEEEexample:BSTcontrol}

\title{Investigating Deep Learning Benchmarks for Electrocardiography Signal Processing \\
}

% 1\textsuperscript{st}

\DeclareRobustCommand*{\IEEEauthorrefmark}[1]{%
  \raisebox{0pt}[0pt][0pt]{\textsuperscript{\footnotesize\ensuremath{#1}}}}

\author{
    \IEEEauthorblockN{
        Hao Wen\IEEEauthorrefmark{1}, Jingsu Kang\IEEEauthorrefmark{2}
    }\\
    \IEEEauthorblockA{
        \IEEEauthorrefmark{1}LMIB, School of Mathematical Sciences, Beihang University, Beijing, 102206, China\\
        \IEEEauthorrefmark{2}Tianjin Medical University, Tianjin, 300203, China\\
        \IEEEauthorrefmark{1}{\small\{wenh06\}@buaa.edu.cn, \{wenh06\}@gmail.com}\\
        \IEEEauthorrefmark{2}{\small\{kangjingsu\}@tmu.edu.cn, \{kjs890223\}@gmail.com}\\
    }
}

\maketitle

\setcounter{footnote}{0}

\begin{abstract}
% almost finished

In recent years, deep learning has witnessed its blossom in the field of Electrocardiography (ECG) processing, outperforming traditional signal processing methods in various tasks, for example, classification, QRS detection, wave delineation. Although many neural architectures have been proposed in the literature, there is a lack of systematic studies and open-source libraries for ECG deep learning.

In this paper, we propose a deep learning framework, named \texttt{torch\_ecg}, which gathers a large number of neural networks, both from literature and novel, for various ECG processing tasks. It establishes a convenient and modular way for automatic building and flexible scaling of the networks, as well as a neat and uniform way of organizing the preprocessing procedures and augmentation techniques for preparing the input data for the models. Besides, \texttt{torch\_ecg} provides benchmark studies using the latest databases, illustrating the principles and pipelines for solving ECG processing tasks and reproducing results from the literature. \texttt{torch\_ecg} offers the ECG research community a powerful tool meeting the growing demand for the application of deep learning techniques.
\end{abstract}


\newcommand{\repourl}{https://github.com/DeepPSP/torch_ecg}

% \newif\ifanonymous

% % \anonymoustrue
% \anonymousfalse

% % can be placed in 'anonymize.tex' and input to the main file
% \makeatletter
%   \ifanonymous
%     \renewcommand{\repourl}{https://anonymous.4open.science/r/CHIL-submission-1/}
%   \fi
% \makeatother

\paragraph*{Data and Code Availability}
% finished
This paper involves the the following 6 databases: CinC2020 \citep{cinc2020,alday2020cinc2020}, CinC2021 \citep{cinc2021}, CPSC2019 \citep{cpsc2019}, CPSC2020 \citep{cai2020cpsc2020}, CPSC2021 \citep{cpsc2021physionet}, and LUDB \citep{physionet2020ludb,kalyakulina2020ludb}, which will be introduced in more details later in this paper. These databases are hosted either by PhysioNet \citep{goldberger2000physionet} or by corresponding challenge organizing committees and are open, freely accessible to all researchers. PhysioNet \citep{goldberger2000physionet} is a large archive hosting a continuously growing number of datasets and software for physiological signal studies, among which ECG perhaps is the most important. Despite existing classical databases in PhysioNet, several challenges are held yearly as special sessions of conferences along with which large novel databases are released, providing the research community with challenging frontier problems.

The code is available at \url{\repourl}.

\section{Introduction}
\label{sec:intro}
% finished, to check and improve

Research on computer-aided medical auxiliary diagnosis algorithms and systems owns a long history, having a large number of practitioners from both academia and industry. It has abundant research areas and many successful real-world applications. Driven by the remarkable success of the application of deep learning in computer vision (CV) \citep{alexnet,resnet,chollet2017xception}, research on medical imagery is the most prominent and attracting \citep{liu2020deep, Chaunzwa2021lung_dl}. Physiological signal processing, a traditional medical research field, among which electrocardiogram (ECG) is the most extensively studied, attracts many researchers of various backgrounds as well.

Although there has been a steady production of inspiring research work \citep{awni2019stanford_ecg,ribeiro2020automatic} in the field of ECG signal processing, there has always been a lack of an ECG deep learning research framework or library, whose existence would largely facilitate the researchers and improve the overall research levels. With the help of torchvision \citep{pytorch}, medical imagery researchers (especially using ordinary images rather than CT images, MRI images, etc.) can build pipelines quickly and verify the validity of the models or even do neural network architecture searching. A similar situation holds for researchers using the HuggingFace Transformer library \citep{wolf-etal-2020-transformers} from the area of medical text processing.

Based on this motivation, we build a library named \texttt{torch\_ecg}, mainly on top of PyTorch \citep{pytorch}, containing readers of popular ECG databases, common preprocessings and augmentations for ECG, and many neural network architectures. The databases include the many of the PhysioNet databases \citep{goldberger2000physionet}, databases from other challenge competitions \citep{cpsc2018,cpsc2019,cai2020cpsc2020,cpsc2021physionet}, etc. Preprocessing like band-pass filtering, detrending, are organized and monitored in a uniform way; and similarly for the augmentations including Z-score normalization, \texttt{mixup} \citep{zhang2018mixup}, etc. \texttt{torch\_ecg} also collects neural network models from previous literature \citep{awni2019stanford_ecg,ribeiro2020automatic,cai2020rpeak_seq_lab_net,yao2020ati_cnn,moskalenko2019deep}, and provides an convenient way to allow the users to alter the network architecture to search for better ones.

The rest of the paper is organized as follows. In Section \ref{sec:related_work}, preceding work on ECG deep learning is briefly reviewed. An overview of the \texttt{torch\_ecg} library is presented in Section \ref{sec:system}. Afterwards, from Section \ref{sec:proproc_aug} to Section \ref{sec:benchmarks}, we dive into details of \texttt{torch\_ecg} from different prospective. We conclude in Section \ref{sec:conclusions_fw} and propose our future work therein.

\section{Related Work}
\label{sec:related_work}
% finished, to check and improve

To the authors' knowledge, the earliest effort of computer-aided automated ECG diagnosis dates back to the 1960s \citep{pipberger1960preparation}. Since then, this research community has been growing steadily. The most famous and fundamental software for ECG processing is the Waveform Database (WFDB) software package hosted at PhysioNet \citep{wfdb, goldberger2000physionet} and other language versions (for example the most widely used python version \citep{wfdb_python}) of it. This software package was initially developed by George Moody and his colleagues at the MIT Laboratory for Computational Physiology and is still actively updated. WFDB and PhysioNet are epoch-making products that have greatly promoted the research progress of ECG processing. Another outstanding library is the BioSPPy toolbox \citep{biosppy} which bundles together various signal processing methods, not only for ECG. We use these two libraries for data I/O (Input/Output) and basic signal processing.

Deep learning has long been ordinary among various machine learning methods, until the last decade when it started occupying an absolute advantage. The situation is similar for the research field of ECG processing. \citep{awni2019stanford_ecg} used a variant of ResNet \citep{resnet} for the classification of ambulatory (single-lead) ECGs, which was claimed to have achieved cardiologist-level. \citep{ribeiro2020automatic} used another variant of ResNet for classifying the standard 12-lead ECGs, while \citep{yao2020ati_cnn} used convolutional recurrent neural network (CRNN) architecture with a VGG-like \citep{vgg} CNN (convolutional neural network) backbone for similar tasks. Other than classification, neural network models also dominate other ECG downstream tasks. For example, \citep{cai2020rpeak_seq_lab_net} used a branched network structure to tackle the problem of QRS detection, which greatly outperformed existing wavelet-based algorithms and the classical Pan-Tompkins algorithm \citep{pan1985pantompkins}. \citep{moskalenko2019deep} used a U-Net model for wave delineation on 12-lead ECGs and validated the model using the Lobachevsky University Electrocardiography Database (LUDB) \citep{physionet2020ludb,kalyakulina2020ludb}.

Although many powerful deep learning models have been proposed, which greatly promoted the research progress on the problem of automated ECG diagnosis systems, there has not yet existed a software package that broadly collects the state-of-the-art (SOTA) neural network models, just as \citep{rw2019timm} does for CV models, for various tasks of ECG processing, as far as the authors know. The effect of data preprocessing and augmentations on model performance are also not thoroughly studied, lacking statistics and comparative studies. Moreover, unlike other research areas where canonical network structures already exist, for example, convolutions of kernel size 3 for CV models, ECG deep learning models do not have widely-enough accepted network structures, hence flexibility for searching for optimized network hyperparameters and architectures is in desperate need. For the above reasons, \texttt{torch\_ecg} mainly includes
\begin{itemize}
\item uniform implementations of data augmenters, which are monitored by a manager class during the model training process. Similarly for preprocessors.
\item flexible and scalable neural network generation from a comprehensive collection of named neural network architectures from existing literature and networks newly developed by the authors, for various ECG downstream tasks.
\item utilities for building training pipelines, and utilities for logging metrics for facilitating network architectures and hyperparameters searching.
\item benchmark case studies and pretrained models for various tasks using large-scale databases.
\end{itemize}
The above will be interpreted in more detail in the following sections.

\section{System Design}
\label{sec:system}
% finished, to check and improve

Unlike the situation of commercial ECG production systems where data are stored in some database like Apache Impala, data of standard ECG datasets are usually saved in the WFDB file format where signals, annotations and metadata are stored in separate files. The WFDB package \citep{wfdb_python} already provides powerful methods handling a single record in its \texttt{io} submodule. We wrap a standard ECG dataset using a Python class, where paths, the list of ECG records, overall statistics, and other status quantities are pre-loaded and maintained. This is a common practice adopted by frameworks in other research field, for example by \texttt{torchtext}, \texttt{torchvision}, etc. Methods for manipulating a single record along with dataset-specific format conversion functions and helper functions, for example, conversion of annotations of atrial fibrillation into lists of intervals or masks, are wrapped therein as well. We call such a class a \texttt{data\_reader}, which is task-independent (or model-independent). Combined with PyTorch's \texttt{Dataset} and \texttt{DataLoader}, one can quickly set up a pipeline feeding data for training a neural network. If not used for training neural networks, a \texttt{data\_reader} can also facilitate developing for example an SVM model.

ECG processing usually includes procedures of signal preprocessing, including resampling, detrending, denoising, normalizing, etc. Each of them is implemented using a Python class, called a ``preprocessor'', sharing a common calling method and a common abstract base class (ABC). All preprocessors are arranged and manipulated by a manager. The manager also accepts user-defined preprocessors inherited from this common abstract base class and can manipulate them in a uniform way as the built-in preprocessors. A preprocess manager can be instantiated using an ordered configuration dictionary where the preprocessors to use and their parameters are specified. It is recommended to be used inside a PyTorch \texttt{Dataset}. Preprocessors operating on mini-batch tensors are also implemented but less recommended for the following reasons. Preprocessors directly deal with raw ECG signals which might be length-varying. Another reason is that many signal processing algorithms have no PyTorch (GPU) implementation, which results in frequent conversions between numpy arrays and PyTorch tensors. More details of the implemented preprocessors will be given in Section \ref{sec:proproc_aug}.

Training neural networks often involves data augmentation techniques. Data augmentation plays the role of data extrapolation, hence has the potential to improve the model performance outside the distribution of the training data. Common data augmentations include adding Gaussian and/or sinusoidal noise, random flip, mixup \citep{zhang2018mixup}, cutmix \citep{yun2019cutmix}, scaling \citep{yao2020ati_cnn} (named as stretch-or-compress), etc. Some are inherited from computer vision techniques, while others are ECG-specific. The augmenters are implemented as child classes of the PyTorch \texttt{Module} without trainable parameters. The \texttt{forward} functions of them accept the same set of arguments so that a manager which is also a child class of the PyTorch \texttt{Module} can manipulate them uniformly. The augmenters are multiple stochastic, i.e. stochastic in the batch dimension so that the samples to be augmented within one mini-batch are randomly chosen; and also stochastic inside each sample, for example, the amplitudes of Gaussian noise added. In Section \ref{sec:proproc_aug}, the augmenters will be discussed in more details.

A serious problem, compared to computer vision or natural language processing, for the ECG processing research community is that currently there are still no commonly accepted neural network structure design principles. For example, for computer vision neural networks, the convolutions typically have kernel size 3, while ECG neural networks in existing literature usually have larger and varying convolution kernel sizes. ECG signals have a sampling frequency while images do not. The kernel sizes should be related to the sampling frequency (each model should have a fixed sampling frequency input) of the model input.

The most extensively studied architectures are variants of ResNet \citep{awni2019stanford_ecg,ribeiro2020automatic}. They both have kernel size 16\footnote{the former paper \citep{awni2019stanford_ecg} does not state the kernel size used explicitly in the paper, but one can find this in the config file in \href{https://github.com/awni/ecg/blob/master/examples/irhythm/}{github.com/awni/ecg/blob/master/examples/irhythm/}}, however, each using ECGs of sampling frequencies 200Hz and 400Hz as input respectively. The kernel sizes of the convolutions in the QRS detection model proposed in \citep{cai2020rpeak_seq_lab_net} have a decay pattern as the convolutions go deeper in the network. All these indicate that ECG neural networks still need solid work for architectural optimization.

For this reason, the implementation of the models (ref. Table \ref{tab:nn}) in \texttt{torch\_ecg} are modular, hence extremely flexible to adjust the network architectures for the users. We extract the common pattern of a model series and make the rest hyperparameters and components configurable. For example, the vanilla ResNet \citep{resnet}, and its variants \citep{xie2016resnext,he2019resnet-d,awni2019stanford_ecg,ribeiro2020automatic,ridnik2021tresnet} (the convolution part, not including the final classifying fully connected layer) all share a common pattern: one stem, plus 4 stages each of which consists of stacked building blocks of basic type or bottleneck type. DenseNet \citep{densenet} has the pattern of stacked dense blocks, within which convolutional layers are densely connected. Keeping this common pattern, other finer details of the networks, including extra attention modules, choices of activation, method and positions for down-sampling in the network, etc. can be altered very easily by just changing one or more items in the model's configuration. A bunch of such small modules that can be integrated into larger networks are implemented in \texttt{torch\_ecg} as well, including squeeze-and-excitation (SE) module \citep{hu2020senet}, global context (GC) module \citep{cao2019GCNet}, conditional random field (CRF) module \citep{laffertyCrf}, etc. \texttt{torch\_ecg} implements the neural networks in its \texttt{models} module and has a set of built-in configuration files, a large part of which come from existing literature, in the \texttt{model\_configs} module.

\begin{table*}[t]
% finished
  \centering
%   \setlength\tabcolsep{5pt} % default value: 6pt
    % \begin{tabular}{|p{0.17\linewidth}|p{0.4\linewidth}|p{0.28\linewidth}|}
    \begin{tabular}{|c|c|c|}
    \hlineB{3.5}
        Network Family & Selected Publications & Highlights \\ \hline \hline
        \multirow{2}{*}{Branched CNN} & \multirow{2}{*}{\citet{cai2020rpeak_seq_lab_net}} & CPSC2019 rank 1$^*$ \\
         & & CinC2021 rank 3$^\dag$ \\ \hline
        VGG & \citet{yao2020ati_cnn} & CinC2020 rank 1$^\ddag$ \\ \hline
        \multirow{4}{*}{ResNet} & \multirow{4}{*}{\citet{awni2019stanford_ecg,ribeiro2020automatic}} & CPSC2018 rank 1 \\
        & & CinC2020 rank 2-4 \\
        & & CinC2021 rank 1-2 \\
        & & CPSC2021 rank 1 \\ \hline
        DenseNet & \multicolumn{1}{c|}{$\backslash$} & CPSC2020 rank 1 \\ \hline
        Inception & \multicolumn{1}{c|}{$\backslash$} & CinC2021 rank 4 \\ \hline
        Xception & \multicolumn{1}{c|}{$\backslash$} & \multicolumn{1}{c|}{$\backslash$} \\ \hline
        MobileNet & \multicolumn{1}{c|}{$\backslash$} & \multicolumn{1}{c|}{$\backslash$} \\ \hline \hline
        U-Net & \citet{moskalenko2019deep} & CPSC2019 rank 2 \\ \hline
        CRNN & \citet{yao2020ati_cnn,cai2020rpeak_seq_lab_net} & CPSC2019 rank 1$^*$ \\ \hline
        RR-LSTM & \citet{faust2018automated,af_detection} & CPSC2021 rank 5$^{**}$ \\ \hlineB{3.5}
    \end{tabular}
  \caption{Neural networks implemented in \texttt{torch\_ecg}. For the challenges CinC2020-2021 \citep{cinc2020,alday2020cinc2020,cinc2021}, CPSC2018-2021 \citep{cpsc2018,cpsc2019,cai2020cpsc2020,cpsc2021physionet}, more information can be found in Section \ref{sec:benchmarks}. Selected publications are not comprehensive lists, only including those directly dealing with ECG processing rather than the papers that initially proposed the network architectures (usually for CV problems). Details of some networks are not disclosed, for example the rank 1 DenseNet solution of CPSC2020, which is developed by a company.\\
  $^\dag$ not fully branched\\
  $^\ddag$ used in conjunction with transformer encoders \citep{vaswani2017attention}\\
  $^*$ Branched CNN is the CNN backbone, CRNN (convolutional recurrent neural network) is the whole architecture\\
  $^{**}$ used in conjunction with CRNN}
  \label{tab:nn}
\end{table*}

As stated previously, convolutions in neural networks for ECG processing usually have large kernels, hence the depths of the networks are usually largely reduced, to prevent them from growing too large in the number of trainable parameters. To the authors' experiences, vanilla networks directly inherited from computer vision, with only 2D operations changed to 1D operation, perform far worse than their wider and shallower variants. For this reason, \texttt{torch\_ecg} constructs in its \texttt{model\_configs} module various wider and shallower variants of the named neural networks (e.g. TResNet \citep{ridnik2021tresnet}, DenseNet \citep{densenet}, etc.)

To facilitate model training, an abstract base class of \texttt{BaseTrainer} where common operations for training a neural network model in most cases, including the setup of the optimizer, controlling of learning rate scheduler, the workflow of one training epoch, etc., are extracted and implemented. Statistics are maintained by this \texttt{BaseTrainer} using a logging manager, which writes to stdout, text files, CSV files as well as tensorboard files. For training a neural network model for a specific ECG task, one only has to inherit this class and implement a few functions including the process of one step (one mini-batch), and the evaluation function for computing and gathering statistics and metrics. The abstraction of \texttt{BaseTrainer} is moderate, retaining a certain degree of flexibility for the users.

A set of \texttt{Outputs} classes are implemented to facilitate the maintenance of outputs of models for ECG downstream tasks (for more details about ECG downstream tasks, ref. Section \ref{sec:downstream}). Each class has a set of required fields to meet the demand of the downstream task this class corresponds to. \texttt{Outputs} classes have additional functions for the computation of a collection of pre-defined metrics when labels are collected along with model outputs during a training process. These metrics, implemented also as Python classes, are re-implementations from existing literature \citep{moskalenko2019deep,cinc2021,cpsc2019} or from \href{https://en.wikipedia.org/wiki/Precision_and_recall}{Wikipedia}\footnote{\href{https://en.wikipedia.org/wiki/Precision_and_recall}{https://en.wikipedia.org/wiki/Precision\_and\_recall}}.

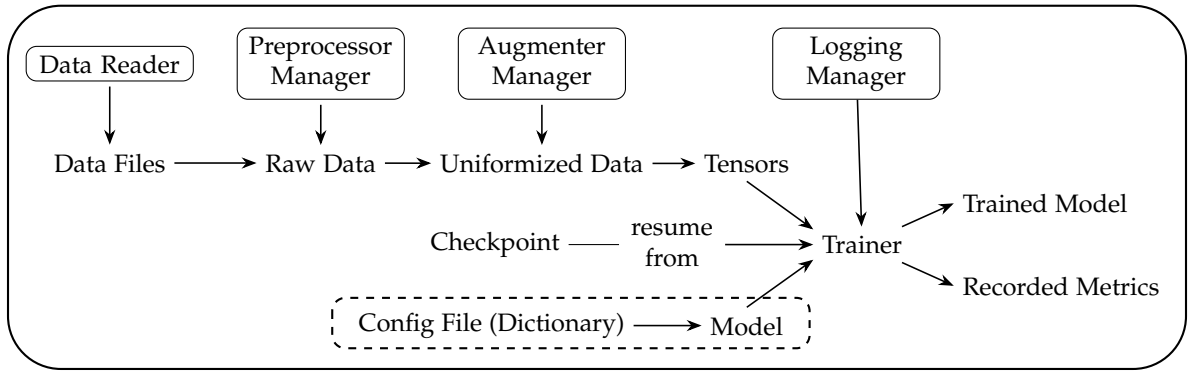
\begin{figure*}[t]
\centering
\begin{tikzpicture}[node distance = 1cm, auto, arr/.style = {semithick, -Stealth}]

\tikzstyle{block} = [rectangle, draw, text width = 6em, text centered, rounded corners, inner sep = 3pt, minimum height = 1.0em]

\node[block] (dr) {Data Reader};
\node[block, right = 1.7em of dr] (pm) {Preprocessor Manager};
\node[block, right = 2.1em of pm] (am) {Augmenter Manager};
\node[below = 2.51em of dr] (df) {Data Files};
\node[below = 1.8em of pm] (rd) {Raw Data};
\node[below = 1.8em of am] (ud) {Uniformized Data};
\node[right = 1.7em of ud] (tensor) {Tensors};

\node[below right = 1.8em and 0.6em of tensor] (trainer) {Trainer};
\node[below = 5em of tensor] (model) {Model};
\node[left = 2.7em of model] (cfg) {Config File (Dictionary)};
\node[block, right = 5.8em of am] (lm) {Logging Manager};
\node[left = 3.5em of trainer, text width = 3.5em, text centered] (resume) {resume from};
\node[left = 2em of resume] (ckpt) {Checkpoint};

\node[above right = 0.2em and 1.7em of trainer] (tm) {Trained Model};
\node[below right = 0.2em and 1.7em of trainer] (rm) {Recorded Metrics};

\draw[arr] ([yshift=-0.2em]dr.south) -- (df);
\draw[arr] ([yshift=-0.2em]pm.south) -- (rd);
\draw[arr] ([yshift=-0.2em]am.south) -- (ud);
\draw[arr] (df) -- (rd);
\draw[arr] (rd) -- (ud);
\draw[arr] (ud) -- (tensor);
\draw[arr] (tensor.south) -- ([yshift=0.6em]trainer.west);
\draw[arr] ([yshift=-0.5em]lm) -- (trainer);
\draw[arr] (cfg) -- (model);
\draw[arr] (model.north) -- ([yshift=-0.6em]trainer.west);
\draw[arr] (trainer) -- (tm.west);
\draw[arr] (trainer) -- (rm.west);
\path[] (ckpt) edge (resume);
\draw[arr] (resume) -- (trainer);

\draw[black,rounded corners=20,thick] ([xshift=-0.7em, yshift=2.3em]dr.west) rectangle ([xshift=0.7em, yshift=-3.3em]rm.east);

\draw[black,rounded corners=5, dashed, thick] ([xshift=-0.7em, yshift=1.1em]cfg.west) rectangle ([xshift=0.7em, yshift=-0.9em]model.east);

\end{tikzpicture}
\caption{Main Modules and Work Flow of \texttt{torch\_ecg}.}
\label{fig:system}
\end{figure*}

The whole system structure and the workflow of \texttt{torch\_ecg} are depicted and summarized in Figure \ref{fig:system}.

\section{Data Preprocessing and Augmentation}
\label{sec:proproc_aug}
% finished, to check and improve

As raw ECG signals may contain noise from various sources, especially those ambulatory ECGs collected from wearable devices, preprocessing sometimes can be crucial for ECG signal processing. One can not guarantee that the training data and the data for inference share similar distributions which are distorted by noises. In extreme cases, even data augmentation can not bridge this gap. Common preprocessing procedures include
\begin{itemize}
    \item Resampling to a fixed sampling rate.
    \item Band-pass filtering, performed using finite impulse response (FIR) filters, Butterworth filters, etc., which removes noises of frequencies outside the given pass band.
    \item Detrending, via median filter, which removes baseline drifts.
    \item Normalization. There are typically 3 ways to do normalization for ECGs as follows
    \begin{equation*}
    \begin{array}{cl}
        \text{Naïve normalization:} & \frac{\mathbf{x} - m}{s} \\
        \text{Z-score normalization:} & \left(\frac{\mathbf{x} - \operatorname{mean}(\mathbf{x})}{\operatorname{std}(\mathbf{x})  + eps}\right) \cdot s + m \\
        \text{Min-Max normalization:} & \frac{\mathbf{x} - \min(\mathbf{x})}{\max(\mathbf{x}) - \min(\mathbf{x}) + eps}
    \end{array}
    \end{equation*}
    where $\mathbf{x}$ is an ECG signal, $m,s$ are given values, a small value $eps$ is added to avoid division by zero error.
\end{itemize}
The naïve normalization method is a standard transformation for images before being fed into a computer vision model, because images have a fixed range of values (0-255 or 0-1), and $m,s$ are usually the mean and standard deviation of the training data respectively. However, it is usually not the case for ECGs, hence naïve normalization is not recommended for preprocessing of ECGs. Z-score normalization is the most commonly used method for ECGs where $m,s$ are typically set $0,1$ respectively. There are researchers \citep{zhang2021over} adopting min-max normalization as well.

As is stated in Section \ref{sec:system}, preprocessors are implemented as child classes of a common abstract base class. The hyperparameters, for example, the cutoff frequencies for the band-pass filter, are accepted as keyword arguments for the instance initialization function \texttt{\_\_init\_\_}. The calling method \texttt{\_\_call\_\_}\footnote{No type hinting and docstring are used here in the code block for simplicity. The same applies to the rest code blocks in this paper. Type hinting is a modern feature of Python that largely improve code readability, hence is strongly recommended to be used.} all accept the signal and sampling frequency as arguments:
\begin{lstlisting}
def __call__(self, sig, fs):
    if len(self.preprocessors) == 0:
        return sig, fs
    ordering = list(range(len(self.preprocessors)))
    if self.random:
        ordering = sample(ordering, len(ordering))
    for idx in ordering:
        sig, fs = self.preprocessors[idx](sig, fs)
    return sig, fs
\end{lstlisting}
In this way, all preprocessors are managed by a \texttt{PreprocManager} whose calling method shares the same set of arguments of the preprocessors.

Data augmentation is an important technique for modern deep learning, and ECG deep learning is no exception. Data augmentation dose extrapolation on the training data, broadening and enriching its distribution. Data augmentation helps alleviate overfitting and over-confidence for neural networks. Most techniques are initially invented for computer vision tasks and are inherited for training ECG neural network models. Augmenters are listed as follows
\begin{itemize}
    \item Addition of Gaussian and/or sinusoidal noise. This technique is sometimes used by researchers and in challenges (e.g. the authors' rank 3 solutions to the problem of supraventricular premature beats (SPB) detection in CPSC2020 \citep{cai2020cpsc2020}), but its efficacy still needs proving and should not be used when preprocessing contains band-pass filtering and/or detrending.
    \item Random Flip, using which the values of the ECGs are multiplied by -1 with a certain probability. This technique also has to be treated carefully whose usage might should be restricted to ambulatory ECGs, since for standard-leads ECGs, altering in relative or absolute values in different leads might completely change its interpretation, for example, the electrical axis.
    \item Mixup \citep{zhang2018mixup}. Mixup uses convex linear combination of two samples for extrapolating the data distribution as follows
    \begin{equation}
    \label{eq:mixup}
    \begin{array}{l}
    \mathbf{x} = \alpha \mathbf{x}_1 + (1-\alpha) \mathbf{x}_2, \\
    \mathbf{y} = \alpha \mathbf{y}_1 + (1-\alpha) \mathbf{y}_2,
    \end{array}
    \end{equation}
    where $(\mathbf{x}_1, \mathbf{y}_1)$, $(\mathbf{x}_2, \mathbf{y}_2)$ are data and labels of 2 samples, $(\mathbf{x}, \mathbf{y})$ is the generated sample, $\alpha \in (0,1)$ is randomly generated from some Beta distribution.
    \item CutMix \citep{yun2019cutmix}. CutMix inserts a part of one sample to another for data distribution extrapolation via
    \begin{equation}
    \label{eq:cutmix}
    \begin{array}{l}
    \mathbf{x} = \mathbf{M} \odot \mathbf{x}_1 + (\mathbf{1}-\mathbf{M}) \odot \mathbf{x}_2, \\
    \mathbf{y} = \alpha \mathbf{y}_1 + (1-\alpha) \mathbf{y}_2,
    \end{array}
    \end{equation}
    the annotations are similar to Equation (\ref{eq:mixup}), $\mathbf{M}$ is a binary mask, in which the proportion of ones is equal to $\alpha$, for cutting and joining two ECGs, $\odot$ is the Hadamard product, i.e. element-wise product. This technique has proven quite useful in the CPSC2021 rank 1 solution.
    \item Masking \citep{yao2020ati_cnn}. This technique randomly masks a small proportion of the signal with zeros, as similar to the Cutout technique \citep{devries2017cutout} in computer vision. This can also be done randomly at critical points, e.g. randomly masking R peaks helps reduce the probability of CNN models to misclassify sinus arrhythmia to atrial fibrillation \citep{af_detection}.
    \item Scaling (stretch-or-compress) \citep{yao2020ati_cnn}. This augmentation method scales the ECG in the time axis, and has proven its validity in \citep{yao2020ati_cnn}.
    \item Label smoothing \citep{inceptionv2v3}. Strictly speaking, label smoothing is not a data augmentation method, but a technique to prevent the model from overconfidence thus improving its capability of generalization. This technique generates soft labels from hard labels in the following way
    \begin{equation}
    \label{eq:label_smooth}
    \mathbf{y}' = (1-\varepsilon) \mathbf{y} + \dfrac{1}{K}\varepsilon \mathbf{1},
    \end{equation}
    where $K$ typically equals the number of classes, $\varepsilon$ is a smoothing factor.
\end{itemize}

All the augmenters mentioned above have successful applications in training ECG neural network models. As is mentioned in Section \ref{sec:system}, they are all implemented as child classes of the Pytorch \texttt{Module}, and accept the same set of parameters for the \texttt{forward} function, hence is manipulated by a \texttt{AugmenterManager} uniformly as follows:

\begin{lstlisting}
def forward(self, sig, label, *extra_tensors, **kwargs):
    if len(self.augmenters) == 0:
        return (sig, label, *extra_tensors)
    ordering = list(range(len(self.augmenters)))
    if self.random:
        ordering = sample(ordering, len(ordering))
    for idx in ordering:
        sig, label, *extra_tensors = \
        self.augmenters[idx](sig, label, *extra_tensors, **kwargs)
    return (sig, label, *extra_tensors)
\end{lstlisting}

\section{Downstream Tasks}
\label{sec:downstream}
% almost finished,

ECG downstream tasks mainly consist of classification and segmentation. Other research topics such as sequence generation (e.g. for denoising), self-supervised representation learning are less active, although having no less importance. Currently, \texttt{torch\_ecg} focuses on the former two tasks, i.e. classification and segmentation.

\begin{figure}[htbp]
\centering
\begin{tikzpicture}[node distance = 1em, auto, arr/.style = {semithick, -Stealth}]
\node[] (cnn) {CNN};
\node[right = 2.1em of cnn.east, text width = 10em] () {ResNet, DenseNet, Branched CNN, etc.};
\node[below = 1.7em of cnn.south] (rnn) {RNN};
\node[right = 2.1em of rnn.east] () {LSTM, etc., optional};
\node[below = 1.7em of rnn.south] (attn) {Attn};
\node[right = 2.1em of attn.east] () {SE, GC, etc., optional};
\node[below = 1.7em of attn.south] (linear) {Linear};

\draw[dashed] ([xshift=-1.4em, yshift=0.3em]rnn.north) rectangle ([xshift=1.4em,yshift=-0.3em]attn.south);

\draw[arr] ([yshift = -0.05em]cnn.south) -- ([yshift = 0.05em]rnn.north);
\draw[arr] ([yshift = -0.05em]rnn.south) -- ([yshift = 0.05em]attn.north);
\draw[arr] ([yshift = -0.05em]attn.south) -- ([yshift = 0.05em]linear.north);

\draw[black,rounded corners=10, thick] ([xshift=-2.1em,yshift=1.1em]cnn.north) rectangle ([xshift=14em,yshift=-0.5em]linear.south);
\end{tikzpicture}
\caption{A typical CRNN architecture. The CNN backbone is the core of this architecture. The optional RNN module and attention module provide some effect of regularization. For classification, feature maps usually go through an additional global pooling layer to cancel the time dimension.}
\label{fig:crnn}
\end{figure}
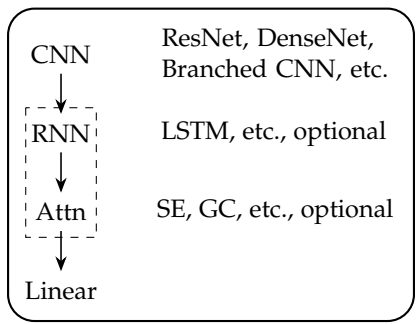

Classification involves producing a probability vector from an ECG signal, usually used for detecting arrhythmias. The most commonly used architectures are CRNN \citep{yao2020ati_cnn} as sketched in Figure \ref{fig:crnn} and its degeneration without recurrent part, i.e. CNN \cite{awni2019stanford_ecg,ribeiro2020automatic}. In most cases, an additional recurrent module slightly improves the model performance at the cost of slowing down the convergence. The CNN backbone is the core of the CRNN architecture, and is the most extensively studied in the literature, as can be inferred from Table \ref{tab:nn}.

Finer-grained tasks including wave delineation (i.e. delineating the P, Q, R, S, T waveforms of the ECG), R peak detection (QRS detection), precise event localization, etc., could be uniformly abstracted as segmentation tasks. This type of tasks can be dealt with using U-Nets \citep{moskalenko2019deep}, fully convolutional networks (FCN), CRNN \citep{cai2020rpeak_seq_lab_net}, etc. When using CRNNs, the difference to a classification model is that the global pooling layer (or layers cancelling the time dimension) is not needed. Instead, one might need upsampling (done by interpolation in PyTorch) to recover the length in the time axis for the model output.

There are also models taking data derived from ECGs as input. For example, in \citep{af_detection, faust2018automated} LSTM models are proposed to predict atrial fibrillation from RR sequences derived from ECGs. \texttt{torch\_ecg} also implements a set of neural networks of such types, allowing additional structures, for example, attention modules, CRF modules, etc., which augment the performance of baseline models \citep{af_detection}.

\section{Benchmark Studies}
\label{sec:benchmarks}
% almost finished

\texttt{torch\_ecg} includes several benchmark studies using databases mentioned in the Data and Code Availability section, to illustrate the usage of this deep learning library, clarifying principles for building pipelines for solving ECG processing tasks, and reproducing results from the literature. The application of \texttt{torch\_ecg} in these benchmark case studies is quite simple. One only needs to \begin{itemize}
\item choose a model type (e.g. CRNN, U-Net, etc.), build a model from a built-in or user-defined configuration file (dictionary), and optionally implement an inference method that includes post-processing of the probability vectors.
\item implement a ``Dataset'' (as a child class of the PyTorch \texttt{Dataset} class) using corresponding ``data\_reader'' from \texttt{torch\_ecg}'s \texttt{databases} module which pre-loads all training data, or loads training data on the fly during training. The implementation of the Dataset depends on the input that the model accepts.
\item inherit \texttt{torch\_ecg}'s abstract base class \texttt{BaseTrainer} to setup a trainer by implementing a function that controls the workflow of one step and a function that evaluates the model, and filling in several mandatory properties, for example the batch dimension of the input, and the monitor for the model selection, etc. The implementation of the workflow of one step is quite simple and in most cases can be of the following form
\begin{lstlisting}
def run_one_step(self, *data):
    signals, labels = data
    signals = signals.to(self.device)
    labels = labels.to(self.device)
    preds = self.model(signals)
    return preds, labels
\end{lstlisting}
The implementation of the evaluation function is slightly more complicated, directly related to the metrics of the ECG tasks.
\end{itemize}

\subsection{PhysioNet/Computing in Cardiology Challenge (CinC) Series}
\label{subsec:cinc}
% almost finished

The CinC series started in 2000 whose topic is detecting sleep apnea from the ECG. In the recent 2 years \citep{cinc2020,alday2020cinc2020,cinc2021}, CinC focuses on the problem of detecting a wide range of ECG abnormalities from the standard 12-lead ECGs and critical problems derived from it. \texttt{torch\_ecg} creates benchmark studies using CinC2020 and CinC2021 challenge databases. The fundamental model architecture used is the CRNN family. Examples are provided using various CNN backbones, including variants from \citep{awni2019stanford_ecg, ribeiro2020automatic} and from \citep{ridnik2021tresnet}, as well as from \citep{cai2020rpeak_seq_lab_net}, etc. Such models have proven their effectiveness in the two challenges. The novelty of these two challenges is their wide range of ECG abnormalities (totally 130+) that are included in the databases. Another innovation point is the study of data redundancy of the ECGs, i.e. using a subset of the standard 12 leads for detecting the ECG abnormalities. \texttt{torch\_ecg} develops a mechanism of ``lead-wise'' convolutions for this problem.

\begin{figure}[htbp]
\centering
\includegraphics[width=\linewidth]{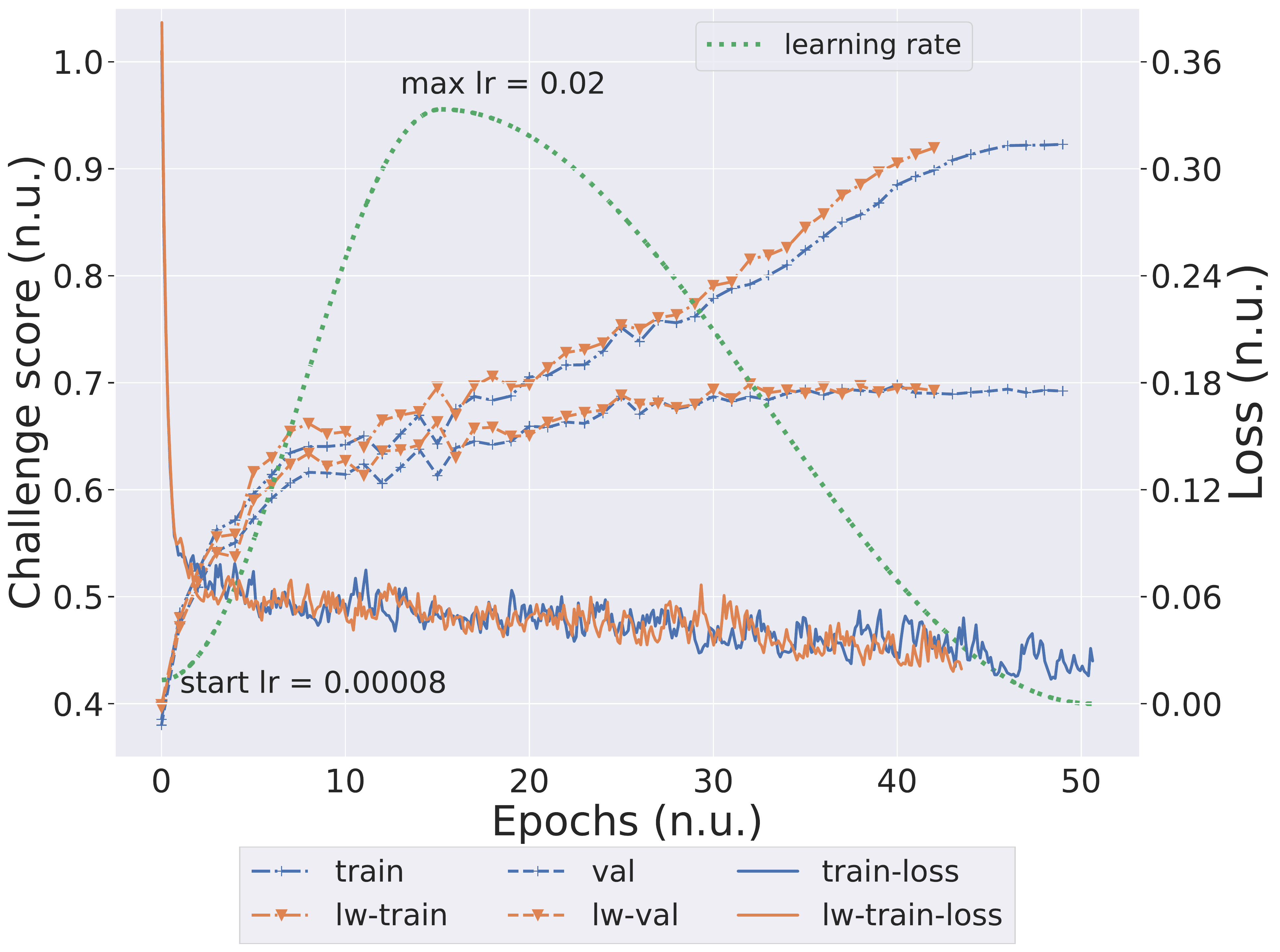}
\caption{Statistics of the training processes of two CRNN models on the CinC2021 database. One uses the CNN backbone proposed in \citep{cai2020rpeak_seq_lab_net}, the other uses its ``lead-wise'' variant (with ``lw'' prefix). The computation of the challenge score can be found in Section 2.2.3 of \citep{alday2020cinc2020}.}
\label{fig:lw-cinc2021}
\end{figure}

Typical experiments comparing a ``normal'' neural network model and its ``lead-wise'' (abbreviated as ``lw'') variant are illustrated in Figure \ref{fig:lw-cinc2021}. The asymmetric loss function \citep{ridnik2021asymmetric_loss} is used for model parameter optimization. The optimizer is the AMSGrad variant of AdamW \citep{adamw_amsgrad}, used in conjunction with the \texttt{OneCycleLR} \citep{smith2019one_cycle} learning rate scheduler of maximum learning rate 0.002 and maximum total epoch number 50. This combination of optimizer and learning rate scheduler will be frequently used in the rest benchmark case studies.

\subsection{China Physiological Signal Challenge (CPSC) Series}
\label{subsec:cpsc}
% almost finished

The CPSC series started in 2018 with the initial problem of the classification of 8 types of ECG abnormalities (totally 9, plus the normal sinus rhythm (NSR)) from the standard 12-lead ECGs, whose database now becomes a part of the CinC2020/2021 challenge database. Afterwards, the problems of QRS detection \citep{cpsc2019}, precise localization of premature beats \citep{cai2020cpsc2020} and atrial fibrillation (AF) events \citep{cpsc2021physionet} are raised. Such problems have finer granularity, as opposed to classification problems, hence are more difficult to solve. \texttt{torch\_ecg} provides multiple or hybrid solutions using various models for these problems. For example, the problem of the detection of AF events raised by CPSC2021 \citep{cpsc2021physionet} requires precise (beat-wise) localization of onsets and offsets of the AF segments in varying-length 2-lead ECGs. It is particularly difficult for paroxysmal AF ECGs where there are multiple AF segments in one ECG recording. CRNN models and U-Net models both apply to this problem, and so does LSTM models using RR intervals derived from the ECGs.

\begin{figure}[htbp]
\centering
\includegraphics[width=\linewidth]{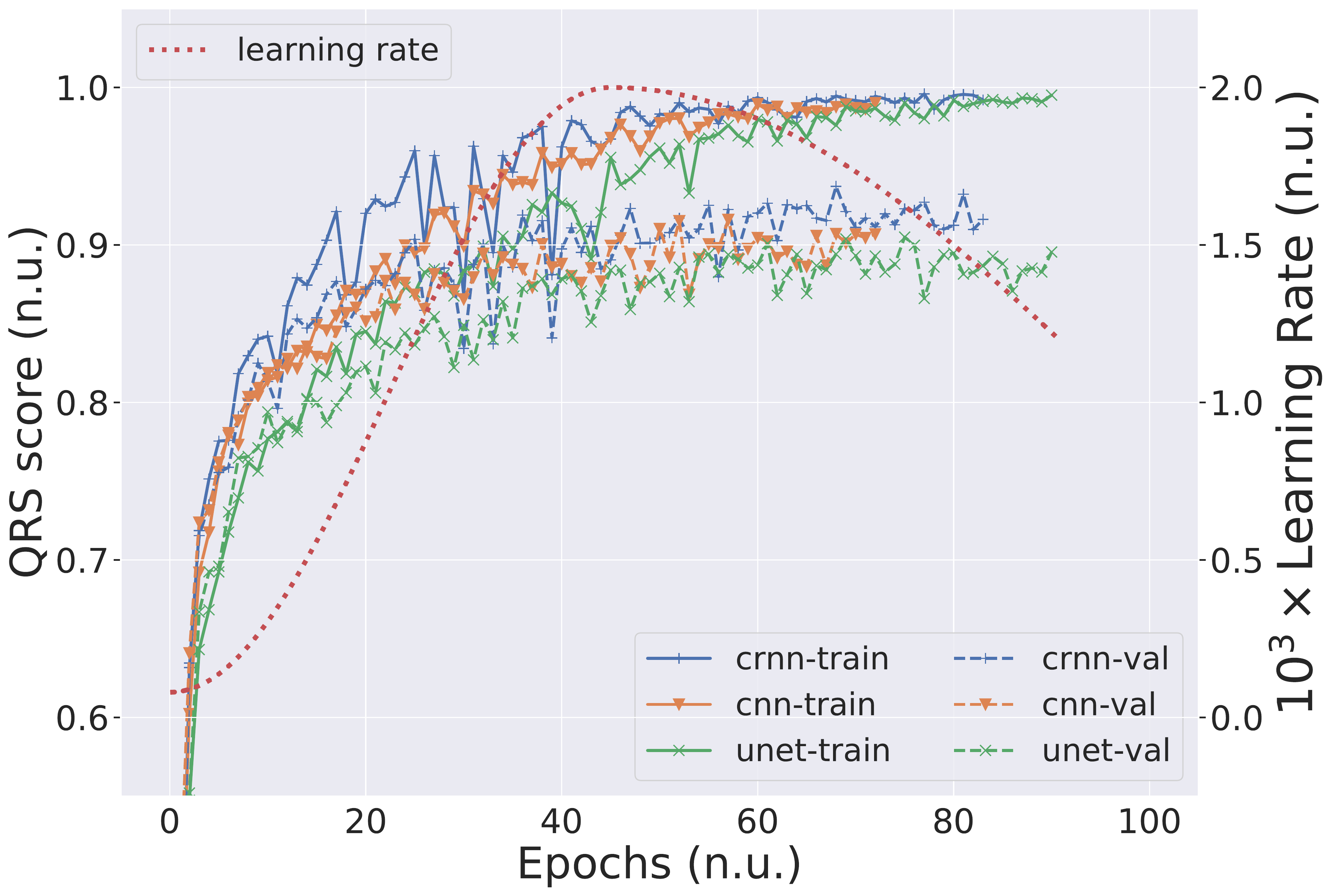}
\caption{Experiments on CPSC2019 using CRNN, pure CNN and U-Net models. Training setups are similar to the experiments in Figure \ref{fig:lw-cinc2021} but with the maximum total epoch number changed to 150. Early stopping is triggered in each of the 3 experiments.}
\label{fig:multi-cpsc2019}
\end{figure}

Results of experiments reproducing \citep{cai2020rpeak_seq_lab_net} along with a experiment using U-Net \citep{moskalenko2019deep} on the benchmark CPSC2019 is shown in Figure \ref{fig:multi-cpsc2019}. The metric ``QRS score'' is defined in Section 3.1 of \citep{cpsc2019}. Note that its original implementation has bugs that score a prediction slightly higher than it should be scored. \texttt{torch\_ecg} re-implements this metric function and fixes the bugs in this benchmark study.

\subsection{Other PhysioNet Datasets}
\label{subsec:physionet}
% almost finished

Despite the classical MIT-BIH Arrhythmia Database \citep{moody2001mitdb}, many new and high-quality databases are proposed in recent years at PhysioNet. The Lobachevsky University Electrocardiography Database (LUDB) \citep{kalyakulina2020ludb,physionet2020ludb} was released in 2020. This database consists of standard 12-lead ECGs, with annotations of the P waves, T waves and the QRS complexes. Along with this database, a U-Net model \citep{moskalenko2019deep} was proposed for the task of ECG wave delineation.

\begin{figure}[htbp]
\centering
\includegraphics[width=\linewidth]{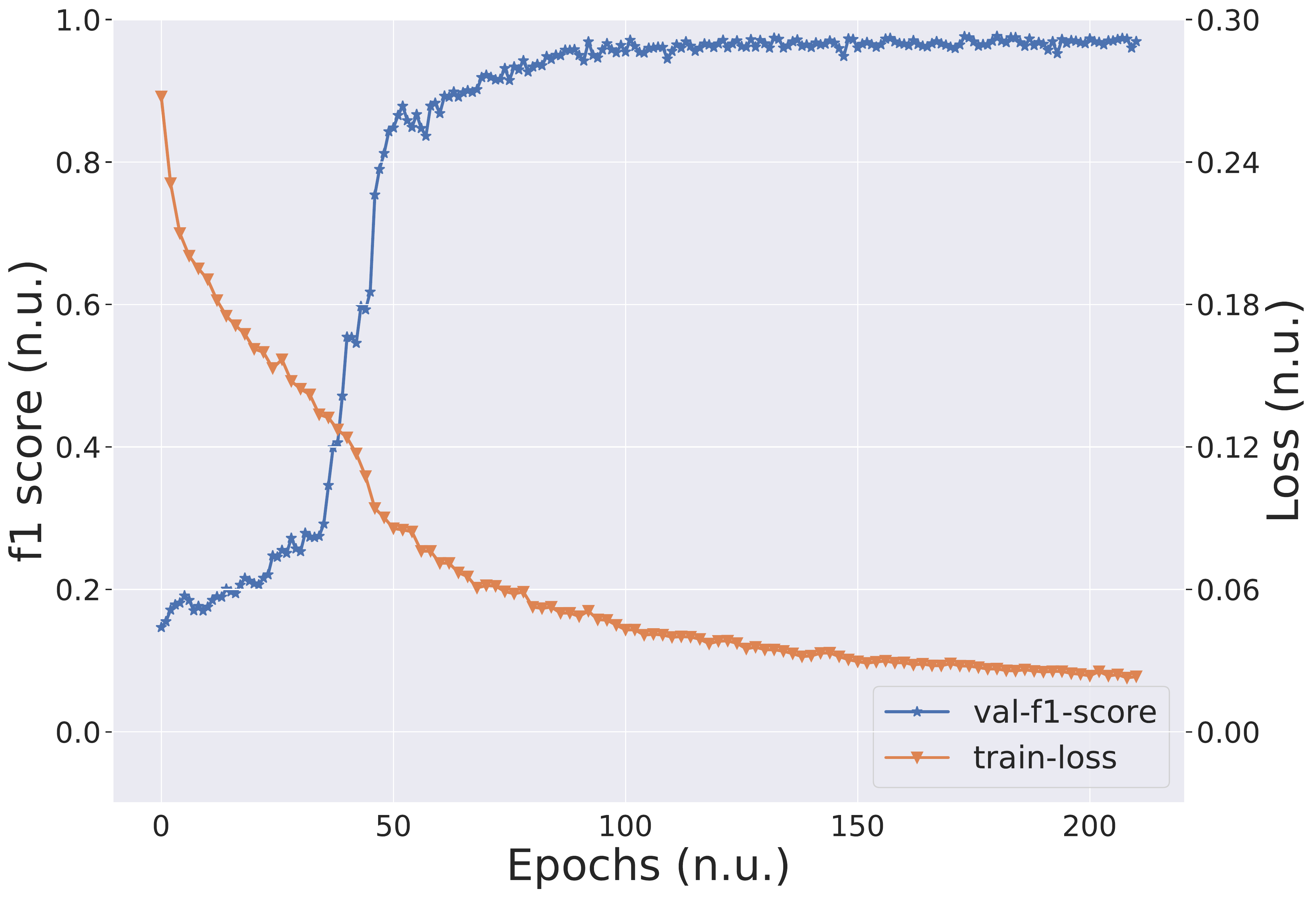}
\caption{A typical experiment of training a U-Net model for wave delineation using the LUDB. The f1 score is the macro-averaged of f1 scores of the 3 waveforms, namely the P wave, the T wave and the QRS complex, as defined in \citep{moskalenko2019deep}. The focal loss function \citep{lin2017focal_loss} is used for model parameter optimization. A burst improvement of the f1 score can be observed near epoch 40.}
\label{fig:unet-ludb}
\end{figure}

\texttt{torch\_ecg} re-implements, with slight adjustments, this work as a benchmark case study. A typical experiment is depicted in Figure \ref{fig:unet-ludb}. Metrics, including sensitivity, precision, mean error, standard deviation and f1 score which is the monitor for model selection, are computed following Section 3.2 of \citep{moskalenko2019deep}. The loss function used is again the focal loss function.

The full list of currently implemented benchmark studies in \texttt{torch\_ecg} is collected in Table \ref{tab:benchmarks}.

\begin{table*}[t]
\centering
\begin{tabular}{|c|c|c|c|c|}
\hlineB{3.5}
Database & ECG type & Annotations & Tasks & Models \\
\hlineB{2}
CinC2020 & standard 12-lead & ECG abnormalities & classification & CRNN \\
CinC2021 & standard 12-lead & ECG abnormalities & classification & CRNN \\
CPSC2019 & single-lead & R peaks & segmentation & CRNN, U-Net \\
CPSC2020 & single-lead & premature beats & segmentation & CRNN \\
CPSC2021 & 2-lead & R peaks, AF episodes & segmentation & CRNN, U-Net, RR-LSTM \\
LUDB & standard 12-lead & ECG waveforms & segmentation & U-Net \\
\hlineB{3.5}
\end{tabular}
\caption{Summary of benchmark studies included in \texttt{torch\_ecg}. Note that beat-wise classification is treated as segmentation in the table. Most annotations are given in the WFDB annotation file format, a few are given by MATLAB files.}
\label{tab:benchmarks}
\end{table*}

\section{Conclusions and Future Work}
\label{sec:conclusions_fw}
% almost finished

\texttt{torch\_ecg} provides the ECG research community with an open-source deep learning library, aiming at bridging the gap between its growing demand for deep learning and the current status that such resources are dispersed. It offers a convenient and modular way for automatic building and flexible scaling of the networks and provides a neat and uniform way of organizing the preprocessing procedures and augmentation techniques for preparing the input data for the models. It gathers and maintains a comprehensive list of neural network architectures both from literature and novel. Moreover, it includes several benchmark case studies as tutorials for using this library, which serve as guidelines and principles for building pipelines for solving a wide range of ECG processing tasks. \texttt{torch\_ecg} would continue to grow and evolve, collecting and creating more powerful deep learning techniques and more effective neural network architectures.

\section*{Institutional Review Board (IRB)}
This work does not require IRB approval.

\section*{Acknowledgments}
% finished

The authors would like to thank professor Deren Han from LMIB, School of Mathematical Sciences, Beihang University and professor Wenjian Yu from the Department of Computer Science and Technology, BNRist, Tsinghua University for generously providing GPU servers to help accomplish this work. This work is supported by NSFC under grant No. 11625105, 12131004.

% \section*{Code Availability}
% % finished

% Code are all available at \newline
% \href{https://github.com/DeepPSP/torch_ecg}{https://github.com/DeepPSP/torch\_ecg}

% \input{playground}

% \bibliographystyle{IEEEtran}
% \bibliography{references}
\printbibliography

% \begin{IEEEbiography}
% [{\includegraphics[width=1in, height=1.25in, clip, keepaspectratio]{bio_images/bio_wenhao.jpg}}]{Hao Wen}
% \end{IEEEbiography}

\end{document}